\documentclass[11pt]{TinLatex} 

\usepackage{graphicx}
\usepackage{epstopdf}

\usepackage{amsmath,amsxtra,amssymb,latexsym, amscd}

\usepackage[mathscr]{eucal}

\usepackage{hyperref}
\hypersetup{
 colorlinks = true,
 linkcolor  = red,
 citecolor  = red,
 urlcolor   = black
}

\usepackage{xspace}

\def\centerbmp#1#2#3{\vskip#2\relax\centerline{\hbox to#1{\special
  {bmp:#3 x=#1, y=#2}\hfil}}}

\def\centereps#1#2#3{\vskip#2\relax\centerline{\hbox to#1{\special
  {eps:#3 x=#1, y=#2}\hfil}}}

\def\centerwmf#1#2#3{\vskip#2\relax\centerline{\hbox to#1{\special
  {wmf:#3 x=#1, y=#2}\hfil}}}

\def\centerps#1#2#3{\vskip#2\relax\centerline{\hbox to#1{\special
  {ps:#3 x=#1, y=#2}\hfil}}}

\input dih.tex

\font\it=cmti12 at 11pt
\font\bf=cmbx12 at 11pt

\advance\hoffset0.4cm
\font\bfh=cmssbx10 scaled 1400
\font\bfl=cmssbx10 scaled \magstep1

\font\chu=cmr8

\font\cn=cmr10

\def\vn{\vskip0.11cm}

\def\ed{\end{document}}
\def\vuong{\raise-0.16cm\hbox{$^\blacksquare$}}

\makeatletter
\DeclareRobustCommand\onedot{\futurelet\@let@token\@onedot}
\def\@onedot{\ifx\@let@token.\else.\null\fi\xspace}

\makeatother

\usepackage{multirow}
\usepackage{enumitem}
\usepackage{algorithm}
\usepackage{amsmath}

\NewDocumentCommand{\MYref}{O{black}mo}{%
  \begingroup
  \hypersetup{linkcolor=#1}%
  \ref{#2}%
  \IfValueT{#3}{%
    \color{#1}{#3}%
  }%
  \endgroup
}

\usepackage{caption} 
\usepackage{cleveref}
\usepackage{color,soul}
\usepackage{pagecolor}
\usepackage{amsthm}
\usepackage{amsfonts}
\usepackage{amssymb}
\usepackage{amsmath}

\usepackage{orcidlink}
\usepackage{algpseudocode}
\usepackage{indentfirst}

\usepackage{booktabs}
\usepackage{tikz}
\usetikzlibrary{patterns}
\usepackage{url}
\usepackage{mathtools}
\usepackage{textcomp}
\usepackage{lscape}

\begin{document}
\makeatletter	   
\setcounter{page}{1}
\renewcommand{\ps@plain}{%
	\renewcommand{\@oddhead}{\hfill\begin{tabular}{r}
    \small{\emph{Journal of Computer Science and Cybernetics, V.xx, N.xx (20xx), 1--}} \\
    \footnotesize{DOI:~10.15625/1813-9663/xx/x/xxxx}
		\end{tabular}
	}

	\renewcommand{\@evenhead}{\@oddhead}
	\renewcommand{\@oddfoot}{\small\hfill{\copyright\ 20xx Vietnam  Academy of Science \& Technology}}
	\renewcommand{\@evenfoot}{\@oddfoot}
}

\def\footnoterule{\kern-3\p@
	\hrule \@width 1.32in \kern 2.6\p@} 
\newcommand{\fn}[1]{\footnotetext{\hspace{-6mm}#1}}
\setlength{\skip\footins}{8mm}

\makeatother   


\title{SOLVING THE TWO-DIMENSIONAL SINGLE STOCK SIZE CUTTING STOCK PROBLEM WITH SAT AND MAXSAT}
\author{
	{\cn TUYEN VAN KIEU$^1$, CHI LINH HOANG$^1$ and KHANH VAN TO$^1$}
	\vskip.5cm
	{
	\it $^1$VNU University of Engineering and Technology, Hanoi, Vietnam \\
\fn{*Corresponding author.}
\fn{\hspace{1.7mm}{\it E-mail addresses}:
\href{mailto:tuyenkv@vnu.edu.vn}{tuyenkv@vnu.edu.vn} (TUYEN VAN KIEU);
\href{mailto:22028132@vnu.edu.vn}{22028132@vnu.edu.vn} (CHI LINH HOANG);
\href{mailto:khanhtv@vnu.edu.vn}{khanhtv@vnu.edu.vn} (KHANH VAN TO).
}}}
\maketitle
\renewcommand\refname{\normalsize \centerline{ REFERENCES}}
\pagestyle{plain}
\pagestyle{myheadings}
\markboth{\footnotesize \chu \uppercase{TUYEN VAN KIEU} {\it et al.}}
{\footnotesize  \chu \uppercase{Solving the 2D-CSSP with SAT and MaxSAT}}

{\cn
\begin{abstract}{Cutting rectangular items from stock sheets to satisfy demands
while minimizing waste is a central manufacturing task. The Two-Dimensional Single
Stock Size Cutting Stock Problem (2D-CSSP) generalizes bin packing by requiring
multiple copies of each item type, which causes a strong combinatorial blow-up.
We present a SAT-based framework where item types are expanded by demand, each copy
has a sheet-assignment variable and non-overlap constraints are activated only for
copies assigned to the same sheet. We also introduce an
\emph{infeasible-orientation elimination} rule that fixes rotation variables when
only one orientation can fit the sheet. For minimizing the number of sheets, we
compare three approaches: non-incremental SAT with binary search, incremental SAT
with clause reuse across iterations and weighted partial MaxSAT. On the Cui--Zhao
benchmark suite, our best SAT configurations certify two to three times more
instances as provably optimal and achieve lower optimality gaps than OR-Tools,
CPLEX and Gurobi. The relative ranking among SAT approaches depends on rotation:
incremental SAT is strongest without rotation, while non-incremental SAT is more
effective when rotation increases formula size.}
\vn
\end{abstract}
}


\section{INTRODUCTION}\label{sec:intro}

In everyday manufacturing (steel plate cutting~\cite{jariyavajee2025efficient},
glass panel production~\cite{hasbiyati2025glass}, furniture board
layout~\cite{reinders1992cutting} and garment fabric
optimization~\cite{farley1988mathematical}), a recurring task is to cut a list of
rectangular items from identical raw sheets while using as few sheets as possible.
When each item type $t$ carries a demand $d_t\ge 1$, the problem is known as the
Two-Dimensional Single Stock Size Cutting Stock Problem
(2D-CSSP)~\cite{gilmore1965multistage,haessler1991cutting}. Because the special case
$d_t=1$ already coincides with Two-Dimensional Bin Packing (2D-BPP), an established
NP-hard problem~\cite{garey1979}, the cutting stock variant inherits that hardness
and adds a new layer of difficulty rooted in the multiplicative effect of demands on
the search space.

As surveyed in Section~\ref{sec:related}, existing approaches each leave a critical
gap. Exact methods (column generation~\cite{gilmore1965multistage}, MIP
models~\cite{furini2013models,chen1995analytical} and enumeration
algorithms~\cite{steyn2015exact}) scale poorly when item demands are large, because
demand multiplies both the pricing-subproblem difficulty and the number of geometric
variables. Heuristic and metaheuristic methods~\cite{coffman1980performance,
cui2013heuristic,he2009hybrid} handle large instances efficiently but provide
\emph{no certificates of optimality}. Even commercial CP and MIP solvers prove
optimality on only a fraction of moderately-sized instances within practical time
limits, as our experiments in Section~\ref{sec:vs_commercial} will demonstrate.

Three specific research gaps emerge from this landscape. First, \emph{no declarative
logic-based formulation} (SAT, MaxSAT, or constraint-satisfaction
encoding) has been proposed for 2D-CSSP. SAT-based methods exist for the related
strip packing~\cite{soh2010} and bin packing~\cite{van2025sat}
problems, but the demand dimension (item-type replication, conditional non-overlap
across sheets and demand-induced symmetries) is unique to the cutting stock setting
and requires a fundamentally different encoding layer. Second, \emph{the interaction
between incremental SAT solving and demand structure has never been
investigated}: copies of the same item type produce geometrically identical
sub-problems, suggesting that conflict clauses should transfer effectively across
solver iterations, yet no study has tested this hypothesis. Third, \emph{no
comprehensive comparison} exists between SAT-based approaches and industrial
optimisation platforms on a common 2D-CSSP benchmark, leaving the relative strengths
of these paradigms unknown.

We argue that Boolean Satisfiability (SAT) is well suited to fill these gaps. The
\emph{order encoding}~\cite{tamura2009} of coordinates represents each integer
variable with a linear number of Booleans, keeping the encoding compact. More
importantly, the conflict-driven clause learning (CDCL) engine of a modern solver
like Glucose~\cite{audemard2009glucose} can derive geometric conflict clauses (e.g.,
that two wide items can never sit side by side on one sheet) and reuse them across
all copies of the same item type, directly exploiting the demand structure that
hinders other exact methods. Incremental SAT solving~\cite{een2003extensible}
amplifies this benefit: when testing whether $k$ sheets suffice, conflict clauses
learned during the (failed) attempt with $k{-}1$ sheets survive and prune the new
search. Weighted partial MaxSAT~\cite{li2021maxsat} additionally allows the sheet
count to be minimised in a single solver invocation. Yet \emph{transferring these
techniques to 2D-CSSP is not straightforward}: the presence of demands requires a new
encoding layer (conditional activation of overlap constraints based on sheet
assignment) and creates additional demand-induced symmetries absent in single-item
packing problems.

The specific contributions of this paper are:
\begin{enumerate}
    \item The \emph{first SAT encoding for 2D-CSSP}, to our knowledge, in which
    item types are replicated by demand, each copy receives a sheet-assignment
    variable and non-overlap constraints are \emph{conditional} on same-sheet
    assignment.
    \item An \emph{infeasible-orientation elimination} rule that fixes the rotation
    variable via a unit clause whenever only one orientation fits the sheet,
    strictly generalising the square-item fix of~\cite{soh2010}.
    \item An \emph{empirical study of three solving approaches} (non-incremental SAT,
    incremental SAT and MaxSAT) on 2D-CSSP, showing that their relative ranking
    differs from 2D-BPP~\cite{van2025sat} and depends on whether rotation is enabled.
    \item A \emph{large-scale comparison} with OR-Tools~9.10, CPLEX~22.1.1 and
    Gurobi~13.0 on the 30-instance Cui--Zhao suite, where our SAT-based approach
    certifies two to three times more optima and achieves a markedly lower gap.
\end{enumerate}

The remainder of this paper is organised as follows. Section~\ref{sec:related} surveys related work on exact, heuristic and SAT-based methods for cutting and packing. Section~\ref{sec:formulation} formalises the 2D-CSSP and derives sheet-count bounds. Section~\ref{sec:encoding} presents the SAT encoding with conditional non-overlap constraints and symmetry-breaking rules. Section~\ref{sec:search} describes the three solving strategies and their trade-offs. Section~\ref{sec:experiments} evaluates all configurations on the Cui--Zhao benchmark suite against four commercial solvers and Section~\ref{sec:conclusion} concludes with future directions.

\section{RELATED WORK}\label{sec:related}

Among exact methods, column generation (CG), originating from Gilmore and
Gomory~\cite{gilmore1965multistage}, is the dominant paradigm for
cutting stock. A master LP selects cutting patterns while a pricing subproblem
generates new ones by solving a two-dimensional knapsack
problem~\cite{viswanathan1993best}. Branch-and-price wraps integer rounding around
this loop. While theoretically elegant, the pricing step is itself NP-hard and its
difficulty grows with item type variety and demands. No publicly available CG
implementation targets the Cui--Zhao benchmark instances used in this work, which
prevents a fair experimental comparison. Integer programming models for two-stage or
multi-stage guillotine cutting~\cite{furini2013models,silva2010integer,
vanderbeck2001nested} achieve exactness but restrict the cutting geometry and their
formulation size grows rapidly with demands. Alternative positional MIP
formulations~\cite{chen1995analytical} generate a prohibitive number of binary
variables when $d_t$ is large; a type with $d_t{=}20$ multiplies the geometric
variables twentyfold. The exact algorithm of Steyn and
Hattingh~\cite{steyn2015exact}, targeted at a related N-sheet variant, relies on
enumeration that scales poorly with the total item count.

\begin{sloppypar}
Because of the scalability barriers above, heuristic and metaheuristic methods are widely used in practice. Level-based
placement~\cite{coffman1980performance}, specialised cutting
rules~\cite{cui2013heuristic} and tabu search~\cite{alvarez2002tabu} handle large
instances efficiently. Hybrid metaheuristics, combining genetic algorithms with
local search~\cite{he2009hybrid} or LP relaxations with tabu
search~\cite{alvarez2002lp}, further improve solution quality. Multi-objective
formulations simultaneously minimise waste and setup cost~\cite{mellouli2019genetic}.
Recent extensions broaden the framework by integrating lot-sizing
decisions~\cite{silva2014integrating}, sequence-dependent setup
times~\cite{wuttke2018sequence} and usable leftovers under demand
uncertainty~\cite{nascimento2023usable}. None of these approaches provides
certificates of optimality.
\end{sloppypar}

More recently, SAT-based and constraint programming methods have proven
competitive for geometric packing. Soh et al.~\cite{soh2010} proposed an
order-encoding SAT formulation for strip packing~\cite{tamura2009}, introducing
large-item and same-size symmetry-breaking rules that we inherit in this work; this
encoding was extended with rotation and modern solver strategies
in~\cite{van2025efficient}. Our prior work~\cite{van2025sat} carried the approach to
the Two-Dimensional Bin Packing Problem (2D-BPP), demonstrating that CDCL learning
and incremental solving offer a strong exact paradigm for packing. The present paper
extends this line of work to 2D-CSSP, where the demand dimension introduces a
qualitatively new encoding challenge (conditional non-overlap constraints and
demand-induced symmetries) that has no analogue in single-item packing.

\section{PROBLEM FORMULATION}\label{sec:formulation}

\subsection{Problem Description}

An instance of 2D-CSSP is a tuple $\langle T, W, H \rangle$ where
$T=\{(w_t,h_t,d_t)\}_{t=1}^{n}$ is a list of item types and $W\times H$ is the
sheet size. Every dimension is a positive integer. The goal is to assign each of the
$N=\sum_t d_t$ required copies to one of $k$ sheets, place them without overlap and
minimize $k$.

Before encoding, we expand the demands: type $t$ yields
copies $c_{t,1},\ldots,c_{t,d_t}$, producing the expanded set $\mathcal{C}$ with
$|\mathcal{C}|=N$. Each copy inherits $w$ and $h$ from its parent type.

For each copy $c\in\mathcal{C}$, we define the following decision variables:
\begin{itemize}
    \item $\sigma_c\in\{1,\ldots,k\}$: assigned sheet index,
    \item $(x_c,y_c)\in\mathbb{Z}_{\ge 0}^{2}$: bottom-left corner,
    \item $r_c\in\{0,1\}$: rotation flag ($r_c=1$: rotated $90^\circ$).
\end{itemize}
Writing $\bar{w}_c = w_c(1{-}r_c)+h_c\,r_c$ and $\bar{h}_c =
h_c(1{-}r_c)+w_c\,r_c$ for the effective dimensions, the feasibility constraints
are:
\begin{align}
  x_c + \bar{w}_c &\le W,\quad y_c + \bar{h}_c \le H, &\forall\,c, \label{eq:boundary}\\
  \sigma_c = \sigma_{c'} &\;\Longrightarrow\; \text{no-overlap}(c,c'), &\forall\,c<c'. \label{eq:cond_overlap}
\end{align}
The no-overlap predicate requires that at least one of the four separating conditions
holds: $c$ left of $c'$, $c'$ left of $c$, $c$ below $c'$, or $c'$ below $c$.
Constraint~\eqref{eq:boundary} enforces in-sheet placement, while
constraint~\eqref{eq:cond_overlap} activates non-overlap only for pairs on the
same sheet.

When $d_t=1$ for every type, the problem recovers 2D-BPP
exactly. A modest 10-type instance with average demand 5 generates $N{=}50$
copies, yielding $\binom{50}{2}=1225$ potential overlap pairs per sheet.

\begin{figure}[ht]
\centering
\begin{tikzpicture}[scale=0.5, every node/.style={font=\scriptsize}]
\node[font=\footnotesize\bfseries, anchor=south] at (2.5, 11) {Item types};
\fill[blue!20] (0,7.5) rectangle (3,9.5);
\draw[blue!60, thick] (0,7.5) rectangle (3,9.5);
\node at (1.5, 8.8) {$t_1{:}\;3{\times}2$};
\node at (1.5, 8.1) {$d_1{=}3$};
\fill[orange!20] (0,4.5) rectangle (2,6.5);
\draw[orange!60, thick] (0,4.5) rectangle (2,6.5);
\node at (1, 5.8) {$t_2{:}\;2{\times}2$};
\node at (1, 5.1) {$d_2{=}3$};
\node at (2.5, 3.5) {$N{=}6$ copies};
\draw[->, line width=1.2pt, gray!70] (5.5, 7) -- (8, 7);
\node[font=\footnotesize\bfseries, anchor=south] at (12, 11) {Optimal packing ($k^*{=}2$)};
\node[font=\footnotesize, anchor=east] at (8.6, 8) {Sheet 1};
\draw[very thick] (9,6) rectangle (15,10);
\fill[blue!20] (9,6) rectangle (12,8);
\draw[blue!60] (9,6) rectangle (12,8);
\node at (10.5,7) {$c_{1,1}$};
\fill[blue!20] (12,6) rectangle (15,8);
\draw[blue!60] (12,6) rectangle (15,8);
\node at (13.5,7) {$c_{1,2}$};
\fill[orange!20] (9,8) rectangle (11,10);
\draw[orange!60] (9,8) rectangle (11,10);
\node at (10,9) {$c_{2,1}$};
\fill[orange!20] (11,8) rectangle (13,10);
\draw[orange!60] (11,8) rectangle (13,10);
\node at (12,9) {$c_{2,2}$};
\fill[orange!20] (13,8) rectangle (15,10);
\draw[orange!60] (13,8) rectangle (15,10);
\node at (14,9) {$c_{2,3}$};
\draw[<->, thick] (9, 5.5) -- (15, 5.5);
\node[below] at (12, 5.4) {$W{=}6$};
\draw[<->, thick] (15.5, 6) -- (15.5, 10);
\node[right] at (15.5, 8) {$H{=}4$};
\node[font=\footnotesize, anchor=east] at (8.6, 2.5) {Sheet 2};
\draw[very thick] (9,0.5) rectangle (15,4.5);
\fill[blue!20] (9,0.5) rectangle (12,2.5);
\draw[blue!60] (9,0.5) rectangle (12,2.5);
\node at (10.5,1.5) {$c_{1,3}$};
\fill[gray!5] (12,0.5) rectangle (15,2.5);
\draw[pattern=north east lines, pattern color=gray!30] (12,0.5) rectangle (15,2.5);
\draw[gray!40] (12,0.5) rectangle (15,2.5);
\fill[gray!5] (9,2.5) rectangle (15,4.5);
\draw[pattern=north east lines, pattern color=gray!30] (9,2.5) rectangle (15,4.5);
\draw[gray!40] (9,2.5) rectangle (15,4.5);
\end{tikzpicture}
\caption{\small A 2D-CSSP instance: sheet size $6{\times}4$, two item types
($t_1{:}\;3{\times}2$, $d_1{=}3$; $t_2{:}\;2{\times}2$, $d_2{=}3$), optimal packing
on $k^*{=}2$ sheets. Hatched regions are waste.}
\label{fig:csp_example}
\end{figure}
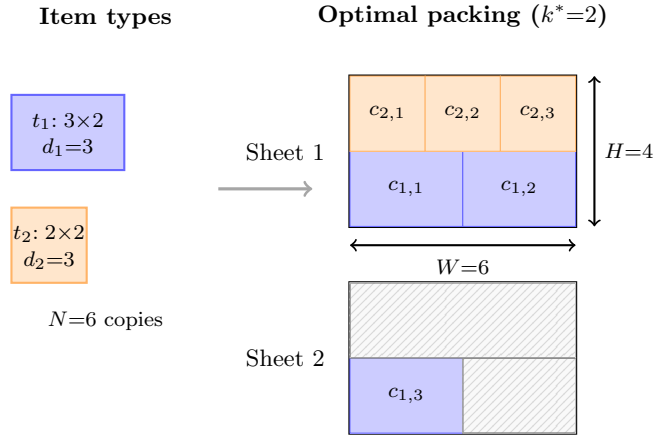

Figure~\ref{fig:csp_example} illustrates a small 2D-CSSP instance.

\subsection{Bounding the Sheet Count}\label{sec:bounds}

Tight bounds on $k$ reduce the number of solver invocations.

For the lower bound, ignoring geometric constraints, $k$ sheets provide total area
$k\!\cdot\!W\!\cdot\!H$, so
\begin{equation}
  k \;\ge\; LB \;=\; \Bigl\lceil \textstyle\sum_{c\in\mathcal{C}} w_c h_c \;\big/\; (W\!\cdot\!H) \Bigr\rceil.
  \label{eq:lb}
\end{equation}
Equation~\eqref{eq:lb} is the area-based lower bound used by all three search
approaches.

For the upper bound, we compute an initial feasible solution using the First-Fit
Decreasing (FFD) heuristic~\cite{coffman1980performance}: sort copies by
non-increasing height, assign each to the first sheet where it fits (shelf-based
layout) and open a new sheet when needed. The number of sheets used is $UB$.

\section{SAT ENCODING}\label{sec:encoding}

Given bounds $LB\le k\le UB$, we build a propositional formula $\Phi_k$ whose models
correspond to feasible packings on $k$ sheets. The encoding consists of order-encoded
coordinates, conditional non-overlap constraints guarded by sheet-assignment
variables and symmetry-breaking rules tailored to the demand setting.

\subsection{Boolean Variables}\label{sec:vars}

\begin{itemize}
  \item \textit{Sheet assignment} $s_{c,j}$: copy $c$ is on sheet $j$
        ($c\!\in\!\mathcal{C},\;j\!\in\!\{1,\ldots,k\}$).
  \item \textit{Position (order encoding)} $\mathit{px}_{c,e}$: $x_c \le e$
        ($e\!\in\!\{0,\ldots,W{-}1\}$); $\mathit{py}_{c,f}$: $y_c \le f$
        ($f\!\in\!\{0,\ldots,H{-}1\}$).
  \item \textit{Relative position} $\ell_{c,c'}$ (``$c$ left of $c'$''),
        $u_{c,c'}$ (``$c$ below $c'$'').
  \item \textit{Rotation} $R_c$: copy $c$ is rotated.
  \item \textit{Sheet usage} $a_j$: at least one copy uses sheet $j$.
\end{itemize}

\subsection{Core Constraints}\label{sec:constraints}

\textit{Sheet assignment (exactly-one).} Every copy goes to exactly one sheet:
\begin{align}
  &\textstyle\bigvee_{j} s_{c,j}, \quad\forall\,c \label{eq:alo}\\
  &\neg s_{c,j_1}\lor \neg s_{c,j_2},\quad\forall\,c\in\mathcal{C},\;\forall\,j_1,j_2\in\{1,\ldots,k\},\;j_1<j_2 \label{eq:amo}
\end{align}
Equations~\eqref{eq:alo}--\eqref{eq:amo} enforce exactly-one sheet assignment per
copy.

\textit{Order encoding axioms.} Position variables must be monotone:
$\mathit{px}_{c,e}\Rightarrow \mathit{px}_{c,e+1}$ and
$\mathit{py}_{c,f}\Rightarrow \mathit{py}_{c,f+1}$.

\textit{Conditional non-overlap.} The key design choice is that non-overlap
constraints are \emph{conditional} on same-sheet assignment. For copies $c,c'$
($c<c'$) and every sheet $j$:
\begin{equation}
  \neg s_{c,j}\lor\neg s_{c',j}\lor \ell_{c,c'}\lor\ell_{c',c}\lor u_{c,c'}\lor u_{c',c}
  \label{eq:sep}
\end{equation}
This six-literal clause differs structurally from the unconditional four-literal
disjunction used in strip-packing encodings~\cite{soh2010}: the two extra guard
literals $\neg s_{c,j},\neg s_{c',j}$ ensure the geometric disjunction is only
enforced when both copies share sheet~$j$.
Constraint~\eqref{eq:sep} is the core conditional non-overlap rule in our CSP
encoding.

\textit{Position-relation consistency.} When $c$ is \emph{not} rotated ($\neg R_c$),
the left-of relation demands $x_c + w_c \le x_{c'}$, encoded as:
\begin{equation}
  \neg s_{c,j}\!\lor\!\neg s_{c',j}\!\lor\! R_c\!\lor\!\neg\ell_{c,c'}\!\lor\!
  \mathit{px}_{c,e}\!\lor\!\neg\mathit{px}_{c',e{+}w_c}
  \label{eq:link_lr}
\end{equation}
for each offset $e \in \{0,\ldots,W{-}w_c{-}1\}$ (with $w_c$ replaced by $h_c$
when $R_c$ is true). Vertical link clauses mirror this structure.
Equation~\eqref{eq:link_lr} links relative-order variables to coordinate variables.

\textit{Domain constraints.} Each copy must fit within its assigned sheet:
\begin{align}
  R_c \lor \mathit{px}_{c,\,W-w_c},\quad &\neg R_c \lor \mathit{px}_{c,\,W-h_c} \label{eq:bnd_x}\\
  R_c \lor \mathit{py}_{c,\,H-h_c},\quad &\neg R_c \lor \mathit{py}_{c,\,H-w_c} \label{eq:bnd_y}
\end{align}
Equations~\eqref{eq:bnd_x} and~\eqref{eq:bnd_y} enforce horizontal and vertical
fit within sheet boundaries.

\textit{Sheet usage.} $\neg s_{c,j}\lor a_j$ for all $c,j$. These variables express
the optimisation objective.

\textit{Encoding correctness.} By construction, every satisfying assignment of
$\Phi_k$ yields a feasible $k$-sheet packing: exactly-one constraints ensure each
copy occupies exactly one sheet, monotonicity axioms guarantee valid order-encoded
coordinates and conditional non-overlap clauses enforce geometric separation for
every pair of copies on the same sheet. Conversely, any feasible $k$-sheet packing
satisfies $\Phi_k$ up to the canonical sheet relabelling imposed by SB4.

\subsection{Symmetry Breaking}\label{sec:sb}

Demand expansion amplifies the symmetry of the search space because the $d_t$ copies
of type $t$ are interchangeable. We deploy four rules.

\textit{(SB1) Large item constraints~\cite{soh2010}.} If $\bar{w}_c+\bar{w}_{c'}>W$
in every orientation combination, the relative position variables
$\ell_{c,c'},\ell_{c',c}$ are set to false (conditioned on same-sheet assignment).
An analogous vertical check is applied.

\textit{(SB2) Same-sized item ordering~\cite{soh2010}.} For two copies $c_i,c_j$ of
the same type ($i<j$), we post $\neg\ell_{c_j,c_i}$: the later copy cannot be placed
strictly to the left of the earlier one. In CSP, this additionally breaks the $d_t!$
permutation symmetry among copies of each type.

\textit{(SB3) Infeasible-orientation elimination.} Soh et
al.~\cite{soh2010} disable rotation for square items
($w_t{=}h_t \Rightarrow \neg R_c$); the same idea is extended
in~\cite{van2025efficient} to force or forbid rotation when an item exceeds the
strip height in one orientation. We generalise further: if only one orientation fits
within the sheet, the rotation variable is fixed via a unit clause. This preserves
soundness by only fixing provably forced orientations.

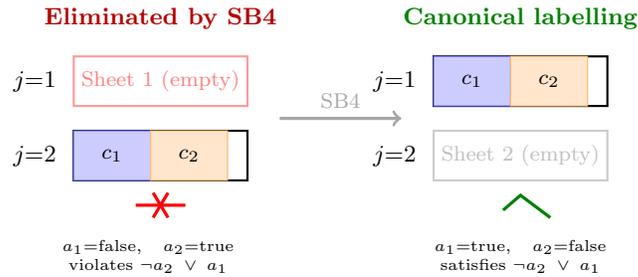
\begin{figure}[ht]
\centering
\begin{tikzpicture}[scale=0.33, every node/.style={font=\scriptsize}]
\node[font=\footnotesize\bfseries, anchor=south, red!70!black] at (3.5, 7.2) {Eliminated by SB4};
\draw[thick, red!40] (0,4.5) rectangle (7,6.5);
\node[red!50] at (3.5, 5.5) {Sheet 1 (empty)};
\draw[thick] (0,1.5) rectangle (7,3.5);
\fill[blue!20] (0,1.5) rectangle (3.1,3.5);
\draw[blue!60] (0,1.5) rectangle (3.1,3.5);
\node at (1.55,2.5) {$c_1$};
\fill[orange!20] (3.1,1.5) rectangle (6.2,3.5);
\draw[orange!60] (3.1,1.5) rectangle (6.2,3.5);
\node at (4.65,2.5) {$c_2$};
\node[font=\footnotesize, anchor=east] at (-0.3, 5.5) {$j{=}1$};
\node[font=\footnotesize, anchor=east] at (-0.3, 2.5) {$j{=}2$};
\draw[red, very thick] (2.5,0.5) -- (4.5,0.5);
\draw[red, very thick] (3.2,0.0) -- (3.8,1.0);
\draw[red, very thick] (3.2,1.0) -- (3.8,0.0);
\draw[->, line width=1.2pt, gray!70] (8.3, 4) -- (13.2, 4);
\node[font=\scriptsize, gray!70, above] at (10.75, 4) {SB4};
\node[font=\footnotesize\bfseries, anchor=south, green!50!black] at (18, 7.2) {Canonical labelling};
\draw[thick] (14.5,4.5) rectangle (21.5,6.5);
\fill[blue!20] (14.5,4.5) rectangle (17.6,6.5);
\draw[blue!60] (14.5,4.5) rectangle (17.6,6.5);
\node at (16.05,5.5) {$c_1$};
\fill[orange!20] (17.6,4.5) rectangle (20.7,6.5);
\draw[orange!60] (17.6,4.5) rectangle (20.7,6.5);
\node at (19.15,5.5) {$c_2$};
\draw[thick, gray!40] (14.5,1.5) rectangle (21.5,3.5);
\node[gray!50] at (18, 2.5) {Sheet 2 (empty)};
\node[font=\footnotesize, anchor=east] at (14.2, 5.5) {$j{=}1$};
\node[font=\footnotesize, anchor=east] at (14.2, 2.5) {$j{=}2$};
\draw[green!50!black, very thick] (17.3,0.3) -- (18,0.9) -- (19.2,0.0);
\node[anchor=north, font=\tiny, text width=4.5cm, align=center] at (3, -0.5)
  {$a_1{=}\text{false},\; a_2{=}\text{true}$\\violates $\neg a_2 \lor a_1$};
\node[anchor=north, font=\tiny, text width=4.5cm, align=center] at (18, -0.5)
  {$a_1{=}\text{true},\; a_2{=}\text{false}$\\satisfies $\neg a_2 \lor a_1$};
\end{tikzpicture}
\caption{\small SB4: the left assignment ($a_1{=}\text{false},a_2{=}\text{true}$)
violates $\neg a_2 \lor a_1$ and is eliminated; the right is the canonical
labelling.}
\label{fig:sb4}
\end{figure}

\textit{(SB4) Sheet ordering.} $\neg a_{j+1}\lor a_j$ for $j=1,\ldots,k{-}1$:
sheets are used in index order, eliminating $k!$ relabelling symmetries.
Figure~\ref{fig:sb4} illustrates this rule.

\section{SOLVING APPROACHES}\label{sec:search}

The encoding of Section~\ref{sec:encoding} decides, for a fixed $k$, whether $k$
sheets suffice. To minimize $k$ we employ three approaches.

\subsection{Non-Incremental SAT with Binary Search}\label{sec:nonincsat}

The simplest approach performs binary search over $k\in[LB,UB]$. At each midpoint
$m$, the formula $\Phi_m$ is generated from scratch and passed to a SAT solver. A SAT
verdict shrinks the upper bound; an UNSAT verdict raises the lower bound. The search
terminates when the bounds meet (Algorithm~\ref{alg:binsat}).

\begin{algorithm}[ht]
\caption{Non-incremental SAT with binary search}\label{alg:binsat}
\textbf{Input:} Expanded copies $\mathcal{C}$, sheet size $W{\times}H$, bounds $LB$, $UB$ \\
\textbf{Output:} Optimal sheet count $k^*$
\begin{algorithmic}[1]
\While{$LB < UB$}
    \State $m \gets \lfloor(LB+UB)/2\rfloor$
    \State Build fresh formula $\Phi_m$ for $m$ sheets
    \If{\textsc{Sat}($\Phi_m$)}
        \State $UB \gets m$
    \Else
        \State $LB \gets m+1$
    \EndIf
\EndWhile
\State \Return $LB$
\end{algorithmic}
\end{algorithm}

Each iteration discards all solver state. This avoids carrying irrelevant learned
clauses from earlier sub-problems, but is wasteful when the gap $UB{-}LB$ is large.

\subsection{Incremental SAT with Binary Search}\label{sec:incsat}

Instead of rebuilding $\Phi_m$ at each step, the incremental approach allocates the
formula for $UB$ sheets once and \emph{disables} surplus sheets through solver
assumptions~\cite{een2003extensible}. To test whether $m$ sheets suffice, the solver
is invoked under the assumption set $A_m=\{\neg a_{m+1},\ldots,\neg a_{UB}\}$,
forcing all sheets beyond $m$ to be inactive. Critically, conflict clauses learned
during an UNSAT proof under $A_m$ remain in the clause database; when the solver
later tests $A_{m+1}$, those clauses immediately prune the search.

This clause reuse is why incremental SAT outperforms the non-incremental variant on
demand-heavy instances: many learned clauses encode geometric impossibilities
(e.g., ``three copies of the widest type cannot share one sheet'') that remain valid
regardless of the target sheet count. Algorithm~\ref{alg:inc} details the procedure.

\begin{algorithm}[ht]
\caption{Incremental SAT with binary search}\label{alg:inc}
\textbf{Input:} $\mathcal{C}$, $W{\times}H$, $LB$, $UB$ \\
\textbf{Output:} Optimal sheet count $k^*$
\begin{algorithmic}[1]
\State Build solver $S$ with $\Phi_{UB}$
\While{$LB < UB$}
    \State $m \gets \lfloor(LB+UB)/2\rfloor$
    \State $A_m\gets\{\neg a_{m+1},\ldots,\neg a_{UB}\}$
    \If{$S.\textsc{SolveUnder}(A_m)=\text{SAT}$}
        \State $UB\gets m$
    \Else
        \State $LB\gets m+1$ \Comment{Conflict clauses retained}
    \EndIf
\EndWhile
\State \Return $LB$
\end{algorithmic}
\end{algorithm}

\subsection{Weighted Partial MaxSAT}\label{sec:maxsat}

\begin{sloppypar}
Both search-based approaches enumerate candidate sheet counts via binary search.
MaxSAT avoids this loop by encoding the objective directly. All constraints of
$\Phi_{UB}$ become \emph{hard clauses}. For each sheet usage variable $a_j$
($j{\ge}LB$) we add the \emph{soft clause} $(\neg a_j)$ with unit weight. A weighted
partial MaxSAT solver then finds an assignment maximising the number of unused sheets,
equivalent to minimizing the sheet count. Because the solver reasons about all
possible sheet counts simultaneously, it can exploit inter-level trade-offs that
binary search cannot see. However, current MaxSAT solvers incur overhead from their
internal lower-bounding mechanisms, making this approach complementary rather than
dominant.
\end{sloppypar}

\section{EXPERIMENTS}\label{sec:experiments}

\subsection{Protocol}\label{sec:protocol}

\textit{Benchmark.} We use 30 instances from the 2D-CSSP suite of Cui and
Zhao~\cite{cui2013heuristic}, which features 7--21 item types per instance, demands
ranging from 1 to 45 and sheet dimensions between $100{\times}100$ and
$4000{\times}3000$. The total copy count $N$ ranges from approximately 30 to over 200.

\textit{Environment.} Google Cloud Platform c3-highmem-8 (8 vCPUs, 64~GB RAM,
Ubuntu 22.04). Time limit: 1,800~s per instance (encoding plus solving).

\textit{SAT/MaxSAT solvers.} Glucose 4.2~\cite{audemard2009glucose} for the
SAT-based approaches; TT-Open-WBO-Inc~\cite{nadel2024ttopenwbo} for MaxSAT.

\begin{sloppypar}
\textit{Baselines.} We compare against three industrial platforms:
Google OR-Tools 9.10~\cite{google2023ortools} (CP-SAT model with \texttt{NoOverlap2D}),
IBM CPLEX 22.1.1~\cite{ibm2023cplex} (both a CP model using \texttt{noOverlap} and a
positional MIP model),
and Gurobi 13.0~\cite{gurobi2023optimizer} (big-$M$ MIP with indicator constraints).
\end{sloppypar}
All solvers ran with default parameters; no instance-specific tuning was applied.
All commercial formulations encode equivalent 2D-CSSP feasibility constraints;
performance differences reflect the solver engine and formulation style rather than
the modelled problem.

\textit{Metrics.} We evaluate each configuration using the following measures:
\begin{itemize}
  \item \textbf{\#Opt}: Number of instances (out of 30) for which the solver
        \emph{certifies} optimality within the time limit.
  \item \textbf{\#Feas}: Number of instances for which the solver finds a solution
        matching the best known solution (BKS) but does \emph{not} prove optimality.
  \item \textbf{Avg~TTB}: Average time-to-best in seconds over the
        \#Opt and \#Feas instances only.
  \item \textbf{Tot.\ Vars} / \textbf{Tot.\ Cls}: Total Boolean variables
        ($\times 10^3$) and total clauses ($\times 10^6$) summed over all
        SAT/MaxSAT instances.
  \item \textbf{Gap (\%)}: Average relative gap to the best known solution,
        $\overline{\mathrm{Gap}}=\frac{1}{30}\sum_{i}\frac{k_i - \mathrm{BKS}_i}{\mathrm{BKS}_i}\times
        100\%$, where $k_i$ is the best sheet count found and $\mathrm{BKS}_i$ is the best
        known feasible value reported by Cui and
        Zhao~\cite{cui2013heuristic}.\footnote{Cui and Zhao employ heuristic methods;
        their values are best known solutions used as reference values. No proven
        lower bounds are available for this benchmark.}
\end{itemize}

\begin{sloppypar}
\textit{Configuration naming.} Each SAT/MaxSAT configuration is labelled as follows:
\emph{CSP} (non-incremental SAT), \emph{CSP\_INC} (incremental SAT), \emph{CSP\_MS}
(MaxSAT); suffix \emph{\_SB} means symmetry-breaking rules SB1--SB4 are enabled;
suffix \emph{\_R} means 90-degree rotation is permitted. For commercial baselines,
\emph{\_SB} indicates an equivalent sheet-ordering constraint was added.
\end{sloppypar}

\subsection{SAT/MaxSAT Performance Without Rotation}\label{sec:sat_norot}

Table~\ref{tab:sat_norot} compares the six SAT/MaxSAT configurations when rotation is
disabled.

\begin{table}[ht]
\centering
\caption{SAT/MaxSAT performance on the Cui--Zhao benchmark (no rotation).
Best values per column are in \textbf{bold}.}
\label{tab:sat_norot}
\setlength{\tabcolsep}{4pt}
\begin{tabular}{@{}lrrrrrr@{}}
\toprule
Config & \#Opt & \#Feas & Avg~TTB (s) & Tot.\ Vars ($\times10^3$) & Tot.\ Cls ($\times10^6$) & Gap (\%) \\
\midrule
\multicolumn{7}{c}{\textit{Non-Incremental SAT with Binary Search}} \\
CSP        & 15 & 3 & 59.1 & \textbf{485.0} & 171.6 & 10.55 \\
CSP\_SB    & 15 & 3 & 56.7 & 485.2 & \textbf{158.2} & 10.78 \\
\addlinespace
\multicolumn{7}{c}{\textit{Incremental SAT with Binary Search}} \\
CSP\_INC     & 15 & 3 & \textbf{42.6} & 489.2 & 209.6 & 10.55 \\
CSP\_INC\_SB & \textbf{16} & 3 & 105.9 & 489.2 & 192.9 & \textbf{9.71} \\
\addlinespace
\multicolumn{7}{c}{\textit{Weighted Partial MaxSAT}} \\
CSP\_MS     & 15 & 0 & 58.0 & 492.4 & 212.2 & 16.68 \\
CSP\_MS\_SB & 15 & 0 & 66.5 & 492.4 & 195.1 & 16.68 \\
\bottomrule
\end{tabular}
\end{table}

\begin{sloppypar}
In the no-rotation setting, incremental SAT with symmetry breaking achieves the
strongest results: CSP\_INC\_SB certifies 16 instances optimal, the highest among all
configurations, with the lowest gap (9.71\%) and 3 additional instances matching the
BKS without proof. This ranking \emph{reverses} the 2D-BPP finding, where
non-incremental SAT is typically superior~\cite{van2025sat}. The reversal is explained
by the demand structure of CSP: copies of the same item type yield geometrically
identical sub-problems, so conflict clauses learned while testing $m$ sheets remain
valid when the solver tests $m{+}1$ sheets. In BPP every item is unique, limiting
such cross-iteration clause transfer.
\end{sloppypar}

Symmetry breaking consistently reduces the total clause count by approximately 8\%
across all three approaches (e.g., from 209.6\,M to 192.9\,M for incremental SAT),
while leaving the variable count virtually unchanged,
as SB1 and SB3 fix Boolean variables at encoding time, triggering unit propagation
that eliminates clauses rather than variables.
For incremental SAT, the pruned symmetric regions help the CDCL engine learn more
general conflict clauses, yielding one additional optimality proof and a lower gap.
For MaxSAT, SB reduces clauses but improves neither the optimality count nor the gap.
MaxSAT underperforms the SAT-based approaches on this benchmark because its internal
lower-bounding mechanism (core-guided or implicit hitting set) incurs substantial
overhead when the number of soft clauses is large. Binary search with SAT reaches the
optimal sheet count in fewer total solving seconds because each individual SAT call is
lightweight compared to the cumulative effort of the MaxSAT engine. This trade-off may
reverse on instances with very tight $UB{-}LB$ gaps, where binary search makes many
iterations.

\subsection{SAT/MaxSAT Performance With Rotation}\label{sec:sat_rot}

Table~\ref{tab:sat_rot} reports results when 90-degree rotation is enabled.

\begin{table}[ht]
\centering
\caption{SAT/MaxSAT performance on the Cui--Zhao benchmark (with rotation).
Best values per column are in \textbf{bold}.}
\label{tab:sat_rot}
\setlength{\tabcolsep}{3pt}
\begin{tabular}{@{}lrrrrrr@{}}
\toprule
Config & \#Opt & \#Feas & Avg~TTB (s) & Tot.\ Vars ($\times10^3$) & Tot.\ Cls ($\times10^6$) & Gap (\%) \\
\midrule
\multicolumn{7}{c}{\textit{Non-Incremental SAT with Binary Search}} \\
CSP\_R       & \textbf{18} & \textbf{4} &  99.2 & \textbf{611.0} & \textbf{525.2} & \textbf{3.24} \\
CSP\_R\_SB   & \textbf{18} & \textbf{4} & 162.0 & 611.4 & 536.2 & \textbf{3.24} \\
\addlinespace
\multicolumn{7}{c}{\textit{Incremental SAT with Binary Search}} \\
CSP\_INC\_R    & \textbf{18} & 2 &  87.7 & 614.2 & 622.2 & 3.77 \\
CSP\_INC\_R\_SB & \textbf{18} & 2 & 133.2 & 614.2 & 609.8 & 3.77 \\
\addlinespace
\multicolumn{7}{c}{\textit{Weighted Partial MaxSAT}} \\
CSP\_MS\_R     & 16 & 0 & \textbf{50.3} & 614.2 & 557.4 & 9.14 \\
CSP\_MS\_R\_SB & 17 & 0 &  60.4 & 614.2 & 609.8 & 8.30 \\
\bottomrule
\end{tabular}
\end{table}

When rotation is enabled, non-incremental and incremental SAT are tied in certified
optima: CSP\_R, CSP\_R\_SB, CSP\_INC\_R and CSP\_INC\_R\_SB all prove 18 instances
optimal. However, CSP\_R finds 4 additional BKS-matching feasible instances with a
3.24\% gap, while CSP\_INC\_R finds only 2 feasible with a 3.77\% gap. Rotation
increases the variable count by about 26\% (from 485\,K to 611\,K for
non-incremental SAT), but the clause count grows approximately threefold
(from 171.6\,M to 525.2\,M), indicating that the combinatorial explosion
resides primarily in clauses. This disproportionate clause growth neutralises the
incremental advantage seen in the compact no-rotation setting.

Comparing Tables~\ref{tab:sat_norot} and~\ref{tab:sat_rot}: rotation improves
solution quality across all approaches: the best optimality count increases from 16
to 18 and the best gap decreases from 9.71\% to 3.24\%, a reduction of more than
three times. Symmetry breaking has no effect on optimality or gap for non-incremental
or incremental SAT in the rotation setting, though it increases CSP\_R's average
time-to-best from 99.2\,s to 162.0\,s without quality gain.
Overall, the choice of SAT approach depends on formula size and demand structure.
Incremental SAT is preferred when the formula is compact (no rotation), because
conflict clauses learned from demand copies transfer effectively across iterations.
When rotation doubles the variable count, the clause database grows large enough
that irrelevant clauses dilute the learned knowledge;
non-incremental SAT avoids this by starting fresh at each midpoint and achieves a
lower gap.

\subsection{Comparison with Commercial Solvers}\label{sec:vs_commercial}

Table~\ref{tab:cross_solver} compares the best SAT configuration from each setting
against the commercial optimisation platforms.

\begin{table}[ht]
\centering
\caption{Best SAT/MaxSAT configurations vs.\ commercial solvers.
Best values per block are in \textbf{bold}.}
\label{tab:cross_solver}
\setlength{\tabcolsep}{5pt}
\begin{tabular}{@{}lrrrr@{}}
\toprule
Config & \#Opt & \#Feas & Avg~TTB (s) & Gap (\%) \\
\midrule
\multicolumn{5}{c}{\textit{No Rotation}} \\
\textbf{CSP\_INC\_SB}   & \textbf{16} & 3 & 105.9 & \textbf{9.71} \\
OR-TOOLS\_CP\_SB         & 7 & 8 & 139.1 & 12.03 \\
GUROBI\_MIP\_SB          & 7 & 0 & 264.3 & 21.09 \\
CPLEX\_MIP\_SB           & 7 & 0 & 264.5 & 21.27 \\
OR-TOOLS\_MIP\_SB        & 5 & 1 & \textbf{46.3} & 22.11 \\
CPLEX\_CP\_SB            & 4 & 1 & 77.8 & 24.51 \\
\midrule
\multicolumn{5}{c}{\textit{With Rotation}} \\
\textbf{CSP\_R}          & \textbf{18} & 4 & 99.2 & \textbf{3.24} \\
OR-TOOLS\_CP\_R\_SB      &  6 & 14 & 200.4 &  4.63 \\
CPLEX\_MIP\_R\_SB        &  6 &  4 & \textbf{16.2} & 13.26 \\
GUROBI\_MIP\_R\_SB       &  6 &  4 &  17.5 & 13.46 \\
OR-TOOLS\_MIP\_R\_SB     &  6 &  3 &  82.8 & 13.36 \\
CPLEX\_CP\_R\_SB         &  1 & 10 &  35.0 & 13.26 \\
\bottomrule
\end{tabular}
\end{table}

The most prominent finding is that SAT-based solvers \emph{certify} optimality on
substantially more instances than all commercial platforms. Without rotation,
CSP\_INC\_SB proves 16 instances optimal, more than double the 7 proved by the best
commercial solvers (OR-TOOLS\_CP\_SB, GUROBI\_MIP\_SB, CPLEX\_MIP\_SB), with a
gap of 9.71\% versus 12.03\% for the best commercial configuration. With rotation,
CSP\_R certifies 18 instances optimal, three times the best commercial count of 6,
with a 3.24\% gap versus 4.63\%.

Among commercial baselines, OR-TOOLS\_CP\_SB stands out as the strongest: it
certifies 7 instances without rotation (gap 12.03\%) and reaches the BKS on 15
instances total (7~proven and 8~feasible). CPLEX\_CP\_SB, by contrast, certifies
only 4 instances (gap 24.51\%), worse than all MIP baselines, despite using the
same \texttt{noOverlap} global constraint. With rotation, the disparity widens
further: CPLEX\_CP\_R\_SB certifies only 1 instance (gap 13.26\%) while
OR-TOOLS\_CP\_R\_SB certifies 6 and achieves a 4.63\% gap. This suggests that
OR-Tools' CP-SAT engine is substantially more effective than CPLEX's CP solver at
exploiting global constraints for this problem class.

A striking observation is that OR-TOOLS\_CP frequently \emph{finds} BKS-matching
solutions but fails to \emph{prove} them optimal. With rotation,
OR-TOOLS\_CP\_R\_SB accumulates 20 total instances (6~proven and 14~feasible),
comparable to SAT's 22 (18~proven and 4~feasible), yet its certified count is
one-third that of SAT. This highlights a fundamental advantage of the SAT approach:
the CDCL proof mechanism provides not only competitive solutions but substantially
stronger optimality certificates.

\begin{sloppypar}
MIP solvers certify 5--7 instances with gaps above 21\% without rotation and
13.26--13.46\% with rotation. They are faster at finding their best solutions
(e.g., 16.2\,s for CPLEX\_MIP\_R\_SB) but prove far fewer instances optimal and
achieve much higher gaps than both SAT and OR-Tools\_CP.
\end{sloppypar}

\section{CONCLUSION}\label{sec:conclusion}

\begin{sloppypar}
We have presented what is, to our knowledge, the first SAT-based framework for
two-dimensional cutting stock optimisation with item demands. The encoding activates
non-overlap constraints only between copies assigned to the same sheet and three
solving approaches (non-incremental SAT, incremental SAT and MaxSAT) offer
complementary strengths depending on the instance profile. On the Cui--Zhao
benchmarks, our best configurations certify two to three times more instances as
provably optimal than all evaluated commercial solvers (16--18 vs.\ 1--7), while also
achieving a markedly lower optimality gap. Among commercial solvers, OR-Tools\_CP
frequently finds BKS-matching solutions but fails to prove them optimal, while
CPLEX\_CP certifies very few instances despite using global constraints,
underscoring the strength of the SAT proof mechanism. The ranking among SAT approaches
depends on formula size: incremental SAT dominates in the compact no-rotation setting
thanks to effective clause transfer across demand copies, while non-incremental SAT
achieves a lower gap and more feasible instances when rotation enlarges the formula.
This interaction has no analogue in the single-item 2D-BPP~\cite{van2025sat} or
strip packing settings. Future directions include tighter lower
bounds, column-generation hybrids where SAT solves the pricing sub-problem and
extension to the heterogeneous stock-size variant. Our source code and benchmark data
are publicly available at \url{https://github.com/cutting-stock/csp}.
\end{sloppypar}



\section*{APPENDIX}\label{app:detailed}

Tables~\ref{tab:detail_norot}
and~\ref{tab:detail_rot} report the per-instance
sheet count~($k$), number of variables (\#Var), number of clauses (\#Cls),
and time-to-best (TTB) for all SAT/MaxSAT configurations on the 30-instance
Cui--Zhao benchmark suite.
Table~\ref{tab:detail_or} reports
the per-instance $k$ and TTB for the commercial/OR solver configurations.
An asterisk ($^*$) marks instances where the solver
certified optimality; a dagger ($^\dagger$) marks instances where the solver
found a solution matching the best known solution (BKS) but did not prove
optimality; unmarked entries timed out without matching the BKS.  A dash (--)
in the TTB column denotes timeout (1\,800\,s).  The number of variables
and clauses are expressed in units of $10^{3}$ and $10^{6}$, respectively.

\begin{landscape}
\begin{table}[p]
\centering
\caption{Per-instance results for SAT/MaxSAT configurations without rotation. $^*$\,=\,certified optimal; $^\dagger$\,=\,feasible matching BKS; unmarked\,=\,timeout. -- denotes timeout (1\,800\,s).}
\label{tab:detail_norot}
\scriptsize
\setlength{\tabcolsep}{2pt}
\begin{tabular}{@{}l r r r r r r r r r r r r r r r r r r r r r r r r@{}}
\toprule
 & \multicolumn{4}{c}{CSP} & \multicolumn{4}{c}{CSP\_SB} & \multicolumn{4}{c}{CSP\_INC} & \multicolumn{4}{c}{CSP\_INC\_SB} & \multicolumn{4}{c}{CSP\_MS} & \multicolumn{4}{c}{CSP\_MS\_SB} \\
\cmidrule(lr){2-5} \cmidrule(lr){6-9} \cmidrule(lr){10-13} \cmidrule(lr){14-17} \cmidrule(lr){18-21} \cmidrule(lr){22-25} 
Instance & $k$ & \#Var & \#Cls & TTB & $k$ & \#Var & \#Cls & TTB & $k$ & \#Var & \#Cls & TTB & $k$ & \#Var & \#Cls & TTB & $k$ & \#Var & \#Cls & TTB & $k$ & \#Var & \#Cls & TTB \\
 &  & ($10^3$) & ($10^6$) & (s) &  & ($10^3$) & ($10^6$) & (s) &  & ($10^3$) & ($10^6$) & (s) &  & ($10^3$) & ($10^6$) & (s) &  & ($10^3$) & ($10^6$) & (s) &  & ($10^3$) & ($10^6$) & (s) \\
\midrule
2 (2) & 2$^*$ & 3.0 & 0.09 & 0.3 & 2$^*$ & 3.0 & 0.08 & 0.6 & 2$^*$ & 3.0 & 0.09 & 0.3 & 2$^*$ & 3.0 & 0.08 & 0.3 & 2$^*$ & 3.0 & 0.09 & 0.2 & 2$^*$ & 3.0 & 0.08 & 0.2 \\
3 (18) & 24 & 12.5 & 4.5 & -- & 24 & 12.5 & 3.5 & -- & 24 & 12.7 & 5.4 & -- & 24 & 12.7 & 4.3 & -- & 24 & 12.9 & 5.6 & -- & 24 & 12.9 & 4.4 & -- \\
A1 (17) & 23 & 12.4 & 4.3 & -- & 23 & 12.4 & 3.2 & -- & 23 & 12.7 & 5.2 & -- & 23 & 12.7 & 3.9 & -- & 23 & 12.8 & 5.4 & -- & 23 & 12.8 & 4.0 & -- \\
A2 (11) & 11$^\dagger$ & 9.8 & 2.0 & 32.2 & 11$^\dagger$ & 9.8 & 1.7 & 29.9 & 11$^\dagger$ & 9.9 & 2.4 & 16.7 & 11$^\dagger$ & 9.9 & 2.0 & 43.9 & 12 & 10.0 & 2.5 & -- & 12 & 10.0 & 2.0 & -- \\
A3 (7) & 8 & 9.0 & 1.4 & -- & 8 & 9.0 & 1.4 & -- & 8 & 9.0 & 1.6 & -- & 8 & 9.0 & 1.6 & -- & 8 & 9.1 & 1.7 & -- & 8 & 9.1 & 1.6 & -- \\
A4 (4) & 5 & 6.4 & 0.5 & -- & 5 & 6.4 & 0.5 & -- & 5 & 6.4 & 0.7 & -- & 4$^*$ & 6.4 & 0.6 & 1083 & 4$^*$ & 6.5 & 0.7 & 46.8 & 4$^*$ & 6.5 & 0.6 & 483 \\
A5 (4) & 4$^*$ & 11.6 & 1.3 & 53.4 & 4$^*$ & 11.6 & 1.3 & 52.7 & 4$^*$ & 11.6 & 1.7 & 34.7 & 4$^*$ & 11.6 & 1.6 & 60.6 & 4$^*$ & 11.7 & 1.7 & 25.4 & 4$^*$ & 11.7 & 1.6 & 19.6 \\
CHL1 (5) & 6 & 18.7 & 3.3 & -- & 6 & 18.7 & 3.2 & -- & 6 & 18.8 & 4.6 & -- & 6 & 18.8 & 4.4 & -- & 7 & 18.9 & 4.7 & -- & 7 & 18.9 & 4.5 & -- \\
CHL2 (3) & 3$^*$ & 2.2 & 0.08 & 0.3 & 3$^*$ & 2.2 & 0.08 & 0.3 & 3$^*$ & 2.2 & 0.08 & 0.3 & 3$^*$ & 2.2 & 0.08 & 0.3 & 3$^*$ & 2.3 & 0.09 & 0.2 & 3$^*$ & 2.3 & 0.08 & 0.2 \\
CHL5 (3) & 3$^*$ & 1.1 & 0.02 & 0.2 & 3$^*$ & 1.1 & 0.02 & 0.2 & 3$^*$ & 1.2 & 0.03 & 0.1 & 3$^*$ & 1.2 & 0.03 & 0.1 & 3$^*$ & 1.2 & 0.04 & 0.1 & 3$^*$ & 1.2 & 0.03 & 0.1 \\
CHL6 (5) & 6 & 21.0 & 4.0 & -- & 6 & 21.0 & 3.9 & -- & 6 & 21.1 & 4.8 & -- & 6 & 21.1 & 4.7 & -- & 6 & 21.2 & 4.8 & -- & 6 & 21.2 & 4.7 & -- \\
CHL7 (6) & 6$^*$ & 26.1 & 6.5 & 58.2 & 6$^*$ & 26.1 & 6.4 & 76.2 & 6$^*$ & 26.2 & 7.6 & 76.2 & 6$^*$ & 26.2 & 7.5 & 57.8 & 6$^*$ & 26.3 & 7.7 & 40.4 & 6$^*$ & 26.3 & 7.6 & 35.4 \\
CU1 (11) & 12 & 26.0 & 10.6 & -- & 12 & 26.0 & 9.8 & -- & 12 & 26.2 & 12.5 & -- & 12 & 26.2 & 11.6 & -- & 13 & 26.3 & 12.7 & -- & 13 & 26.3 & 11.7 & -- \\
CU2 (14) & 15 & 35.6 & 22.9 & -- & 16 & 35.6 & 21.1 & -- & 15 & 35.8 & 26.2 & -- & 15 & 35.8 & 24.1 & -- & 16 & 36.0 & 26.5 & -- & 16 & 36.0 & 24.3 & -- \\
CW1 (9) & 10 & 19.5 & 6.0 & -- & 10 & 19.5 & 5.7 & -- & 10 & 19.7 & 8.0 & -- & 10 & 19.7 & 7.6 & -- & 12 & 19.9 & 8.1 & -- & 12 & 19.9 & 7.7 & -- \\
CW2 (12) & 12$^\dagger$ & 20.1 & 7.9 & 188 & 12$^\dagger$ & 20.1 & 6.9 & 206 & 12$^\dagger$ & 20.3 & 10.8 & 115 & 12$^\dagger$ & 20.3 & 9.4 & 299 & 15 & 20.5 & 10.9 & -- & 15 & 20.5 & 9.5 & -- \\
CW3 (15) & 17 & 47.6 & 42.4 & -- & 17 & 47.6 & 38.0 & -- & 17 & 47.9 & 50.4 & -- & 17 & 47.9 & 45.2 & -- & 19 & 48.1 & 50.7 & -- & 19 & 48.1 & 45.5 & -- \\
Hchl2 (6) & 6$^*$ & 26.0 & 6.5 & 79.2 & 6$^*$ & 26.1 & 6.4 & 79.1 & 6$^*$ & 26.1 & 7.5 & 66.8 & 6$^*$ & 26.1 & 7.4 & 62.3 & 6$^*$ & 26.3 & 7.6 & 36.4 & 6$^*$ & 26.3 & 7.5 & 36.2 \\
Hchl3s (3) & 3$^*$ & 14.2 & 1.4 & 13.2 & 3$^*$ & 14.2 & 1.3 & 13.8 & 3$^*$ & 14.2 & 1.8 & 10.3 & 3$^*$ & 14.2 & 1.7 & 9.9 & 3$^*$ & 14.3 & 1.8 & 6.9 & 3$^*$ & 14.3 & 1.7 & 6.0 \\
Hchl4s (2) & 2$^*$ & 7.6 & 0.4 & 3.6 & 2$^*$ & 7.6 & 0.3 & 3.5 & 2$^*$ & 7.6 & 0.5 & 2.8 & 2$^*$ & 7.6 & 0.5 & 2.7 & 2$^*$ & 7.7 & 0.5 & 1.8 & 2$^*$ & 7.7 & 0.5 & 1.4 \\
Hchl6s (5) & 5$^*$ & 29.4 & 6.5 & 63.8 & 5$^*$ & 29.4 & 6.4 & 63.1 & 5$^*$ & 29.5 & 7.8 & 53.6 & 5$^*$ & 29.5 & 7.7 & 49.3 & 5$^*$ & 29.6 & 7.9 & 30.6 & 5$^*$ & 29.6 & 7.8 & 28.4 \\
Hchl7s (7) & 7$^*$ & 50.1 & 20.9 & 290 & 7$^*$ & 50.1 & 20.7 & 249 & 7$^*$ & 50.2 & 23.9 & 242 & 7$^*$ & 50.2 & 23.6 & 197 & 8 & 50.4 & 24.0 & -- & 8 & 50.4 & 23.7 & -- \\
Hchl8s (1) & 2 & 1.6 & 0.02 & -- & 2 & 1.6 & 0.02 & -- & 2 & 1.6 & 0.03 & -- & 2 & 1.6 & 0.03 & -- & 2 & 1.7 & 0.04 & -- & 2 & 1.7 & 0.03 & -- \\
Hchl9 (10) & 10$^*$ & 19.4 & 5.5 & 121 & 10$^*$ & 19.4 & 5.3 & 112 & 10$^*$ & 19.5 & 6.5 & 66.5 & 10$^*$ & 19.5 & 6.4 & 66.3 & 10$^*$ & 19.7 & 6.7 & 649 & 10$^*$ & 19.7 & 6.5 & 355 \\
HH (2) & 2$^*$ & 3.7 & 0.1 & 0.4 & 2$^*$ & 3.7 & 0.1 & 0.4 & 2$^*$ & 3.7 & 0.1 & 0.6 & 2$^*$ & 3.7 & 0.1 & 0.5 & 2$^*$ & 3.7 & 0.1 & 0.2 & 2$^*$ & 3.7 & 0.1 & 0.2 \\
OF1 (3) & 3$^*$ & 2.7 & 0.1 & 5.7 & 3$^*$ & 2.7 & 0.10 & 1.3 & 3$^*$ & 2.7 & 0.1 & 2.7 & 3$^*$ & 2.7 & 0.1 & 1.7 & 3$^*$ & 2.8 & 0.2 & 10.2 & 3$^*$ & 2.8 & 0.1 & 10.5 \\
OF2 (4) & 4$^*$ & 2.9 & 0.2 & 1.0 & 4$^*$ & 2.9 & 0.1 & 1.7 & 4$^*$ & 2.9 & 0.2 & 0.7 & 4$^*$ & 2.9 & 0.2 & 0.6 & 4$^*$ & 3.0 & 0.2 & 0.5 & 4$^*$ & 3.0 & 0.2 & 0.5 \\
STS2 (12) & 12$^\dagger$ & 20.0 & 6.1 & 136 & 12$^\dagger$ & 20.0 & 5.9 & 108 & 12$^\dagger$ & 20.6 & 10.0 & 65.2 & 12$^\dagger$ & 20.6 & 9.6 & 64.8 & 18 & 20.7 & 10.2 & -- & 18 & 20.7 & 9.8 & -- \\
STS4 (5) & 5$^*$ & 12.3 & 1.8 & 16.8 & 5$^*$ & 12.3 & 1.7 & 22.6 & 5$^*$ & 12.3 & 2.1 & 12.3 & 5$^*$ & 12.3 & 2.1 & 11.1 & 5$^*$ & 12.4 & 2.1 & 20.7 & 5$^*$ & 12.4 & 2.1 & 20.9 \\
W (18) & 23 & 12.4 & 4.3 & -- & 23 & 12.4 & 3.1 & -- & 23 & 13.1 & 6.8 & -- & 23 & 13.1 & 4.9 & -- & 30 & 13.3 & 7.0 & -- & 30 & 13.3 & 5.0 & -- \\
\bottomrule
\end{tabular}
\end{table}
\end{landscape}

\begin{landscape}
\begin{table}[p]
\centering
\caption{Per-instance results for SAT/MaxSAT configurations with rotation. $^*$\,=\,certified optimal; $^\dagger$\,=\,feasible matching BKS; unmarked\,=\,timeout. -- denotes timeout (1\,800\,s).}
\label{tab:detail_rot}
\scriptsize
\setlength{\tabcolsep}{2pt}
\begin{tabular}{@{}l r r r r r r r r r r r r r r r r r r r r r r r r@{}}
\toprule
 & \multicolumn{4}{c}{CSP\_R} & \multicolumn{4}{c}{CSP\_R\_SB} & \multicolumn{4}{c}{CSP\_INC\_R} & \multicolumn{4}{c}{CSP\_INC\_R\_SB} & \multicolumn{4}{c}{CSP\_MS\_R} & \multicolumn{4}{c}{CSP\_MS\_R\_SB} \\
\cmidrule(lr){2-5} \cmidrule(lr){6-9} \cmidrule(lr){10-13} \cmidrule(lr){14-17} \cmidrule(lr){18-21} \cmidrule(lr){22-25} 
Instance & $k$ & \#Var & \#Cls & TTB & $k$ & \#Var & \#Cls & TTB & $k$ & \#Var & \#Cls & TTB & $k$ & \#Var & \#Cls & TTB & $k$ & \#Var & \#Cls & TTB & $k$ & \#Var & \#Cls & TTB \\
 &  & ($10^3$) & ($10^6$) & (s) &  & ($10^3$) & ($10^6$) & (s) &  & ($10^3$) & ($10^6$) & (s) &  & ($10^3$) & ($10^6$) & (s) &  & ($10^3$) & ($10^6$) & (s) &  & ($10^3$) & ($10^6$) & (s) \\
\midrule
2 (2) & 2$^*$ & 3.6 & 0.2 & 0.7 & 2$^*$ & 3.6 & 0.2 & 0.8 & 2$^*$ & 3.6 & 0.2 & 0.8 & 2$^*$ & 3.6 & 0.2 & 0.7 & 2$^*$ & 3.6 & 0.2 & 0.3 & 2$^*$ & 3.6 & 0.2 & 0.5 \\
3 (18) & 19 & 15.6 & 15.1 & -- & 19 & 15.6 & 15.2 & -- & 19 & 15.6 & 16.1 & -- & 19 & 15.6 & 12.6 & -- & 19 & 15.6 & 16.1 & -- & 19 & 15.6 & 12.6 & -- \\
A1 (17) & 17$^\dagger$ & 15.4 & 13.5 & 126 & 17$^\dagger$ & 15.5 & 13.6 & 187 & 18 & 15.8 & 18.6 & -- & 18 & 15.8 & 16.6 & -- & 22 & 15.8 & 18.6 & -- & 22 & 15.8 & 16.6 & -- \\
A2 (11) & 11$^\dagger$ & 12.5 & 6.7 & 83.6 & 11$^\dagger$ & 12.5 & 6.7 & 82.9 & 11$^\dagger$ & 12.6 & 8.1 & 42.6 & 11$^\dagger$ & 12.6 & 8.1 & 34.8 & 12 & 12.6 & 8.1 & -- & 12 & 12.6 & 8.1 & -- \\
A3 (7) & 7$^*$ & 11.4 & 4.4 & 42.2 & 7$^*$ & 11.4 & 4.4 & 37.0 & 7$^*$ & 11.5 & 5.0 & 34.3 & 7$^*$ & 11.5 & 5.0 & 24.8 & 7$^*$ & 11.5 & 5.0 & 89.8 & 7$^*$ & 11.5 & 5.0 & 84.0 \\
A4 (4) & 4$^*$ & 8.2 & 1.5 & 264 & 4$^*$ & 8.2 & 1.6 & 177 & 4$^*$ & 8.2 & 1.9 & 37.2 & 4$^*$ & 8.2 & 1.9 & 183 & 5 & 8.2 & 1.9 & -- & 4$^*$ & 8.2 & 1.9 & 276 \\
A5 (4) & 4$^*$ & 14.6 & 3.7 & 34.3 & 4$^*$ & 14.6 & 3.7 & 43.5 & 4$^*$ & 14.7 & 4.7 & 61.2 & 4$^*$ & 14.7 & 4.6 & 29.0 & 4$^*$ & 14.7 & 4.7 & 56.9 & 4$^*$ & 14.7 & 4.6 & 46.7 \\
CHL1 (5) & 6 & 22.8 & 9.1 & -- & 6 & 22.8 & 9.2 & -- & 6 & 22.9 & 11.0 & -- & 6 & 22.9 & 11.0 & -- & 6 & 22.9 & 11.0 & -- & 6 & 22.9 & 11.0 & -- \\
CHL2 (3) & 3$^*$ & 3.0 & 0.2 & 0.7 & 3$^*$ & 3.0 & 0.2 & 0.8 & 3$^*$ & 3.0 & 0.2 & 0.8 & 3$^*$ & 3.0 & 0.2 & 0.7 & 3$^*$ & 3.0 & 0.2 & 0.5 & 3$^*$ & 3.0 & 0.2 & 0.6 \\
CHL5 (3) & 3$^*$ & 1.4 & 0.08 & 0.5 & 3$^*$ & 1.4 & 0.08 & 0.6 & 3$^*$ & 1.4 & 0.1 & 0.3 & 3$^*$ & 1.4 & 0.1 & 0.3 & 3$^*$ & 1.4 & 0.1 & 0.4 & 3$^*$ & 1.4 & 0.1 & 0.3 \\
CHL6 (5) & 6 & 25.6 & 10.9 & -- & 6 & 25.6 & 10.9 & -- & 6 & 25.7 & 13.1 & -- & 6 & 25.7 & 13.1 & -- & 6 & 25.7 & 13.1 & -- & 6 & 25.7 & 13.1 & -- \\
CHL7 (6) & 6$^*$ & 31.1 & 17.4 & 56.4 & 6$^*$ & 31.1 & 17.5 & 61.6 & 6$^*$ & 31.1 & 17.5 & 89.8 & 6$^*$ & 31.1 & 17.5 & 66.4 & 6$^*$ & 31.1 & 17.5 & 61.5 & 6$^*$ & 31.1 & 17.5 & 63.3 \\
CU1 (11) & 12 & 32.7 & 33.1 & -- & 12 & 32.7 & 33.2 & -- & 12 & 32.9 & 39.2 & -- & 12 & 32.9 & 39.2 & -- & 13 & 32.9 & 39.2 & -- & 13 & 32.9 & 39.2 & -- \\
CU2 (14) & 15 & 46.6 & 73.2 & -- & 15 & 46.6 & 73.5 & -- & 16 & 46.8 & 84.0 & -- & 16 & 46.8 & 83.2 & -- & 16 & 46.8 & 84.0 & -- & 16 & 46.8 & 83.2 & -- \\
CW1 (9) & 10 & 24.9 & 18.4 & -- & 10 & 24.9 & 18.5 & -- & 10 & 25.1 & 22.6 & -- & 10 & 25.1 & 22.5 & -- & 11 & 25.1 & 22.6 & -- & 11 & 25.1 & 22.5 & -- \\
CW2 (12) & 12$^\dagger$ & 28.1 & 26.8 & 534 & 12$^\dagger$ & 28.1 & 26.9 & 1623 & 12$^\dagger$ & 28.3 & 34.3 & 603 & 12$^\dagger$ & 28.3 & 33.7 & 1630 & 14 & 28.3 & 34.3 & -- & 14 & 28.3 & 33.7 & -- \\
CW3 (15) & 17 & 65.4 & 139 & -- & 17 & 65.5 & 148 & -- & 17 & 65.7 & 165 & -- & 17 & 65.7 & 165 & -- & 19 & 65.7 & 101 & -- & 19 & 65.7 & 165 & -- \\
Hchl2 (6) & 6$^*$ & 31.1 & 17.4 & 166 & 6$^*$ & 31.1 & 17.5 & 241 & 6$^*$ & 31.2 & 20.4 & 257 & 6$^*$ & 31.2 & 20.3 & 190 & 6$^*$ & 31.2 & 20.4 & 50.8 & 6$^*$ & 31.2 & 20.3 & 52.0 \\
Hchl3s (3) & 3$^*$ & 16.8 & 3.5 & 11.7 & 3$^*$ & 16.8 & 3.5 & 13.8 & 3$^*$ & 16.8 & 3.5 & 13.6 & 3$^*$ & 16.8 & 3.5 & 12.0 & 3$^*$ & 16.8 & 3.5 & 8.4 & 3$^*$ & 16.8 & 3.5 & 9.0 \\
Hchl4s (2) & 2$^*$ & 9.3 & 0.9 & 3.3 & 2$^*$ & 9.3 & 0.9 & 3.9 & 2$^*$ & 9.3 & 0.9 & 3.5 & 2$^*$ & 9.3 & 0.9 & 3.3 & 2$^*$ & 9.3 & 0.9 & 2.1 & 2$^*$ & 9.3 & 0.9 & 2.2 \\
Hchl6s (5) & 5$^*$ & 37.3 & 17.7 & 61.5 & 5$^*$ & 37.3 & 17.7 & 81.7 & 5$^*$ & 37.3 & 17.7 & 88.8 & 5$^*$ & 37.3 & 17.7 & 67.3 & 5$^*$ & 37.3 & 17.7 & 43.1 & 5$^*$ & 37.3 & 17.7 & 43.2 \\
Hchl7s (7) & 7$^*$ & 62.1 & 56.7 & 459 & 7$^*$ & 62.1 & 56.8 & 531 & 7$^*$ & 62.2 & 65.0 & 366 & 7$^*$ & 62.2 & 65.0 & 297 & 8 & 62.2 & 65.0 & -- & 8 & 62.2 & 65.0 & -- \\
Hchl8s (1) & 1$^*$ & 1.9 & 0.04 & 0.6 & 1$^*$ & 1.9 & 0.04 & 0.7 & 1$^*$ & 1.9 & 0.09 & 0.5 & 1$^*$ & 1.9 & 0.09 & 0.3 & 1$^*$ & 1.9 & 0.09 & 30.5 & 1$^*$ & 1.9 & 0.09 & 22.0 \\
Hchl9 (10) & 10$^*$ & 23.0 & 16.2 & 125 & 10$^*$ & 23.0 & 16.3 & 199 & 10$^*$ & 23.1 & 19.5 & 114 & 10$^*$ & 23.1 & 19.5 & 96.2 & 10$^*$ & 23.1 & 19.5 & 439 & 10$^*$ & 23.1 & 19.5 & 398 \\
HH (2) & 2$^*$ & 4.7 & 0.3 & 0.9 & 2$^*$ & 4.7 & 0.3 & 1.0 & 2$^*$ & 4.7 & 0.3 & 1.0 & 2$^*$ & 4.7 & 0.3 & 0.9 & 2$^*$ & 4.7 & 0.3 & 0.6 & 2$^*$ & 4.7 & 0.3 & 0.6 \\
OF1 (3) & 3$^*$ & 3.6 & 0.3 & 2.1 & 3$^*$ & 3.6 & 0.3 & 4.6 & 3$^*$ & 3.7 & 0.5 & 2.9 & 3$^*$ & 3.7 & 0.5 & 1.7 & 3$^*$ & 3.7 & 0.5 & 1.5 & 3$^*$ & 3.7 & 0.5 & 9.0 \\
OF2 (4) & 4$^*$ & 3.9 & 0.5 & 1.3 & 4$^*$ & 3.9 & 0.5 & 1.5 & 4$^*$ & 3.9 & 0.5 & 1.4 & 4$^*$ & 3.9 & 0.5 & 1.3 & 4$^*$ & 3.9 & 0.5 & 1.1 & 4$^*$ & 3.9 & 0.5 & 1.1 \\
STS2 (12) & 12$^\dagger$ & 23.9 & 18.7 & 179 & 12$^\dagger$ & 23.9 & 18.7 & 237 & 13 & 24.2 & 25.5 & -- & 13 & 24.2 & 25.1 & -- & 15 & 24.2 & 25.5 & -- & 15 & 24.2 & 25.1 & -- \\
STS4 (5) & 5$^*$ & 15.1 & 4.9 & 29.8 & 5$^*$ & 15.1 & 4.9 & 35.3 & 5$^*$ & 15.2 & 6.9 & 35.0 & 5$^*$ & 15.2 & 6.9 & 26.0 & 5$^*$ & 15.2 & 6.9 & 17.6 & 5$^*$ & 15.2 & 6.9 & 18.4 \\
W (18) & 20 & 15.6 & 15.1 & -- & 20 & 15.6 & 15.2 & -- & 19 & 15.9 & 19.5 & -- & 19 & 15.9 & 15.3 & -- & 23 & 15.9 & 19.5 & -- & 23 & 15.9 & 15.3 & -- \\
\bottomrule
\end{tabular}
\end{table}
\end{landscape}

\begin{landscape}
\begin{table}[p]
\centering
\caption{Per-instance results for commercial/OR solver configurations. $^*$\,=\,certified optimal; $^\dagger$\,=\,feasible matching BKS; unmarked\,=\,timeout. -- denotes timeout (1\,800\,s).}
\label{tab:detail_or}
\scriptsize
\setlength{\tabcolsep}{2pt}
\begin{tabular}{@{}l r r r r r r r r r r r r r r r r r r r r@{}}
\toprule
 & \multicolumn{4}{c}{\shortstack{CPLEX\\CP}} & \multicolumn{4}{c}{\shortstack{CPLEX\\MIP}} & \multicolumn{4}{c}{\shortstack{GUROBI\\MIP}} & \multicolumn{4}{c}{\shortstack{OR-TOOLS\\CP}} & \multicolumn{4}{c}{\shortstack{OR-TOOLS\\MIP}} \\
\cmidrule(lr){2-5} \cmidrule(lr){6-9} \cmidrule(lr){10-13} \cmidrule(lr){14-17} \cmidrule(lr){18-21} 
 & \multicolumn{2}{c}{w/o R} & \multicolumn{2}{c}{w/ R} & \multicolumn{2}{c}{w/o R} & \multicolumn{2}{c}{w/ R} & \multicolumn{2}{c}{w/o R} & \multicolumn{2}{c}{w/ R} & \multicolumn{2}{c}{w/o R} & \multicolumn{2}{c}{w/ R} & \multicolumn{2}{c}{w/o R} & \multicolumn{2}{c}{w/ R} \\
\cmidrule(lr){2-3} \cmidrule(lr){4-5} \cmidrule(lr){6-7} \cmidrule(lr){8-9} \cmidrule(lr){10-11} \cmidrule(lr){12-13} \cmidrule(lr){14-15} \cmidrule(lr){16-17} \cmidrule(lr){18-19} \cmidrule(lr){20-21} 
Instance & $k$ & TTB & $k$ & TTB & $k$ & TTB & $k$ & TTB & $k$ & TTB & $k$ & TTB & $k$ & TTB & $k$ & TTB & $k$ & TTB & $k$ & TTB \\
 &  & (s) &  & (s) &  & (s) &  & (s) &  & (s) &  & (s) &  & (s) &  & (s) &  & (s) &  & (s) \\
\midrule
2 (2) & 2$^*$ & 4.1 & 2$^\dagger$ & 0.0 & 2$^*$ & 2.1 & 2$^*$ & 2.6 & 2$^*$ & 0.9 & 2$^*$ & 0.4 & 2$^*$ & 0.3 & 2$^*$ & 10.3 & 2$^*$ & 4.1 & 2$^*$ & 20.3 \\
3 (18) & 24 & -- & 19 & -- & 24 & -- & 19 & -- & 23 & -- & 19 & -- & 23 & -- & 19 & -- & 24 & -- & 19 & -- \\
A1 (17) & 23 & -- & 22 & -- & 23 & -- & 22 & -- & 23 & -- & 23 & -- & 23 & -- & 17$^\dagger$ & 107 & 23 & -- & 18 & -- \\
A2 (11) & 12 & -- & 12 & -- & 12 & -- & 12 & -- & 12 & -- & 12 & -- & 11$^\dagger$ & 26.3 & 11$^\dagger$ & 32.5 & 12 & -- & 12 & -- \\
A3 (7) & 8 & -- & 8 & -- & 8 & -- & 8 & -- & 8 & -- & 8 & -- & 8 & -- & 7$^\dagger$ & 30.3 & 8 & -- & 8 & -- \\
A4 (4) & 5 & -- & 5 & -- & 5 & -- & 5 & -- & 5 & -- & 5 & -- & 5 & -- & 4$^\dagger$ & 355 & 5 & -- & 5 & -- \\
A5 (4) & 5 & -- & 4$^\dagger$ & 0.6 & 5 & -- & 5 & -- & 5 & -- & 5 & -- & 4$^\dagger$ & 22.8 & 4$^\dagger$ & 26.4 & 5 & -- & 5 & -- \\
CHL1 (5) & 7 & -- & 6 & -- & 7 & -- & 6 & -- & 7 & -- & 6 & -- & 6 & -- & 6 & -- & 7 & -- & 6 & -- \\
CHL2 (3) & 3$^\dagger$ & 0.0 & 3$^\dagger$ & 0.0 & 3$^*$ & 286 & 3$^\dagger$ & 0.5 & 3$^*$ & 286 & 3$^\dagger$ & 0.3 & 3$^*$ & 7.7 & 3$^\dagger$ & 0.2 & 3$^\dagger$ & 0.1 & 3$^\dagger$ & 0.1 \\
CHL5 (3) & 3$^*$ & 0.2 & 3$^*$ & 376 & 3$^*$ & 0.4 & 3$^*$ & 1.2 & 3$^*$ & 0.4 & 3$^*$ & 1.2 & 3$^*$ & 0.2 & 3$^*$ & 0.3 & 3$^*$ & 1.1 & 3$^*$ & 18.6 \\
CHL6 (5) & 6 & -- & 6 & -- & 6 & -- & 6 & -- & 6 & -- & 6 & -- & 6 & -- & 6 & -- & 6 & -- & 6 & -- \\
CHL7 (6) & 7 & -- & 6$^\dagger$ & 1.8 & 7 & -- & 6$^\dagger$ & 5.0 & 7 & -- & 6$^\dagger$ & 5.0 & 7 & -- & 6$^\dagger$ & 20.3 & 7 & -- & 6$^\dagger$ & 2.3 \\
CU1 (11) & 13 & -- & 13 & -- & 13 & -- & 13 & -- & 13 & -- & 13 & -- & 13 & -- & 13 & -- & 13 & -- & 14 & -- \\
CU2 (14) & 16 & -- & 16 & -- & 16 & -- & 16 & -- & 16 & -- & 16 & -- & 15 & -- & 16 & -- & 16 & -- & 16 & -- \\
CW1 (9) & 12 & -- & 11 & -- & 12 & -- & 12 & -- & 12 & -- & 12 & -- & 10 & -- & 10 & -- & 12 & -- & 11 & -- \\
CW2 (12) & 15 & -- & 13 & -- & 15 & -- & 14 & -- & 15 & -- & 14 & -- & 12$^\dagger$ & 934 & 12$^\dagger$ & 144 & 15 & -- & 14 & -- \\
CW3 (15) & 19 & -- & 18 & -- & 19 & -- & 18 & -- & 19 & -- & 18 & -- & 17 & -- & 18 & -- & 19 & -- & 19 & -- \\
Hchl2 (6) & 7 & -- & 6$^\dagger$ & 1.8 & 7 & -- & 7 & -- & 7 & -- & 7 & -- & 6$^\dagger$ & 265 & 6$^\dagger$ & 1289 & 7 & -- & 7 & -- \\
Hchl3s (3) & 4 & -- & 3$^\dagger$ & 0.5 & 4 & -- & 3$^\dagger$ & 1.2 & 4 & -- & 3$^\dagger$ & 1.2 & 3$^\dagger$ & 30.5 & 3$^\dagger$ & 7.1 & 4 & -- & 4 & -- \\
Hchl4s (2) & 2$^*$ & 321 & 2$^\dagger$ & 0.2 & 2$^*$ & 13.6 & 2$^*$ & 9.2 & 2$^*$ & 13.6 & 2$^*$ & 0.9 & 2$^*$ & 2.2 & 2$^*$ & 296 & 2$^*$ & 82.4 & 2$^*$ & 28.2 \\
Hchl6s (5) & 6 & -- & 5$^\dagger$ & 4.1 & 6 & -- & 6 & -- & 6 & -- & 6 & -- & 5$^\dagger$ & 68.2 & 5$^\dagger$ & 50.4 & 6 & -- & 6 & -- \\
Hchl7s (7) & 8 & -- & 8 & -- & 8 & -- & 8 & -- & 8 & -- & 8 & -- & 8 & -- & 8 & -- & 8 & -- & 8 & -- \\
Hchl8s (1) & 2 & -- & 2 & -- & 2 & -- & 1$^*$ & 89.4 & 2 & -- & 1$^*$ & 158 & 2 & -- & 1$^*$ & 0.6 & 2 & -- & 1$^*$ & 540 \\
Hchl9 (10) & 12 & -- & 11 & -- & 12 & -- & 12 & -- & 12 & -- & 12 & -- & 11 & -- & 11 & -- & 12 & -- & 12 & -- \\
HH (2) & 2$^*$ & 63.1 & 2$^\dagger$ & 0.1 & 2$^*$ & 0.4 & 2$^*$ & 0.8 & 2$^*$ & 0.4 & 2$^*$ & 0.2 & 2$^*$ & 5.3 & 2$^\dagger$ & 0.1 & 2$^*$ & 0.9 & 2$^*$ & 11.6 \\
OF1 (3) & 4 & -- & 4 & -- & 3$^*$ & 711 & 3$^*$ & 51.1 & 3$^*$ & 711 & 3$^*$ & 7.2 & 3$^*$ & 2.2 & 3$^*$ & 1.1 & 3$^*$ & 189 & 3$^*$ & 124 \\
OF2 (4) & 5 & -- & 4$^\dagger$ & 0.0 & 4$^*$ & 838 & 4$^\dagger$ & 1.0 & 4$^*$ & 838 & 4$^\dagger$ & 0.2 & 4$^*$ & 1.2 & 4$^*$ & 606 & 5 & -- & 4$^\dagger$ & 0.2 \\
STS2 (12) & 18 & -- & 13 & -- & 18 & -- & 15 & -- & 18 & -- & 15 & -- & 12$^\dagger$ & 679 & 12$^\dagger$ & 997 & 18 & -- & 16 & -- \\
STS4 (5) & 6 & -- & 6 & -- & 6 & -- & 7 & -- & 6 & -- & 7 & -- & 5$^\dagger$ & 42.4 & 5$^\dagger$ & 34.5 & 6 & -- & 6 & -- \\
W (18) & 30 & -- & 19 & -- & 23 & -- & 20 & -- & 23 & -- & 20 & -- & 23 & -- & 19 & -- & 23 & -- & 20 & -- \\
\bottomrule
\end{tabular}
\end{table}
\end{landscape}

{\small
\bibliographystyle{IEEEtranS}
\bibliography{references}
}

\end{document}